\documentclass[11pt,letterpaper]{article}
\usepackage{emnlp2016}
\usepackage{times}
\usepackage{latexsym}

\emnlpfinalcopy

\def\emnlppaperid{190}

\usepackage[usenames,dvipsnames]{color}
\usepackage{bbding}
\usepackage{graphicx}
\usepackage{multirow}
\usepackage{booktabs}
\usepackage{setspace}
\newcommand{\uedin}{{\sc PBSY}}
\newcommand{\hdu}{{\sc HPB}} %HIER
\newcommand{\kit}{{\sc SPB}} %PBMT
\newcommand{\su}{{\sc NMT}}

\newcommand{\COMMENT}[1] {}
\newcommand{\AB}[1]{\textcolor{black}{#1}}
\newcommand{\checkRight}{\textcolor{OliveGreen}{\Checkmark}}
\newcommand{\checkWrong}{\textcolor{Red}{\XSolidBrush}}

\title{Neural \textit{versus} Phrase-Based Machine Translation Quality: a Case Study}

\author{Luisa Bentivogli  \\FBK, Trento\\Italy \And Arianna Bisazza \\University of Amsterdam\\The Netherlands \And Mauro Cettolo  \\FBK, Trento\\Italy \And Marcello Federico  \\FBK, Trento\\Italy}

\date{}

\begin{document}

\maketitle

\begin{abstract}

Within the field of Statistical Machine Translation (SMT), the neural approach (NMT) has recently emerged as the first
technology able to challenge the long-standing dominance of
phrase-based approaches (PBMT). 
% ORIGINAL: Neural machine translation (NMT) has recently emerged as the first technology able to challenge the long-standing dominance of statistical machine translation (SMT).  
%
In particular, at the IWSLT
2015 evaluation campaign, NMT outperformed well established
state-of-the-art PBMT systems on English-German, a language pair known
to be particularly hard because of morphology and syntactic
differences.
%In this paper, we set out to understand in what respects NMT provides
%better translation quality than SMT.  To this end, 
To understand in what respects NMT provides better translation quality than PBMT,
we perform a detailed analysis of neural \textit{vs.} phrase-based SMT outputs, leveraging
high quality post-edits performed by professional translators on the
IWSLT data.  
%For the first time, our analysis provides useful insights on what linguistic phenomena are best modeled by NMT, while pointing out some other aspects that remain to be improved.
For the first time, our analysis provides useful insights on what
linguistic phenomena are best modeled by neural models -- such as the reordering of verbs --
while pointing out other aspects that remain to be improved.

%for example,
%we will see that NMT output contains less word reordering errors than
%PBMT (-50\%), with a quality peak in the placement of verbs (-70\%
%errors). On the other hand, we will show that some aspects of NMT
%should be improved, like the placement of articles.

\end{abstract}

\section{Introduction}

The wave of neural  models has eventually reached the field of Statistical Machine Translation (SMT). After a  period in which Neural MT (NMT) was too computationally costly and resource demanding to compete with  state-of-the-art Phrase-Based MT (PBMT)\footnote{We use the generic term phrase-based MT to cover standard phrase-based, hierarchical and syntax-based SMT approaches.}, 
the situation changed in 2015. 
For the first time, in the latest edition of IWSLT\footnote{International Workshop on Spoken Language Translation (http://workshop2015.iwslt.org/)} % is a yearly workshop associated with an open evaluation campaign on spoken language translation.} 
\cite{Cettolo:2015},  %a shared evaluation task was dominated by a NMT architecture. 
%On a difficult language pair like English-German, the system described in \cite{luong2015stanford} overtook a variety of PBMT approaches with a large margin ($+5.3$ BLEU points) anticipating what, most likely, will be the new NMT era.
the system described in \cite{luong2015stanford} overtook a variety of PBMT approaches with a large margin ($+5.3$ BLEU points) on a difficult language pair like English-German -- anticipating what, most likely, will be the new NMT era.
% CAMERA-READY new-paragraph

This impressive improvement follows the distance reduction previously observed in the WMT 2015 %\footnote{http://www.statmt.org/wmt15/index.html} 
shared translation task \cite{WMT-15}. Just few months earlier, the NMT systems described in \cite{WMT-15-montreal} ranked on par with the best phrase-based models on a couple of language pairs. Such rapid progress stems from the improvement of the recurrent neural network encoder-decoder model, originally proposed in \cite{Sutskever14,cho14}, with the use of %{\em long short term memory} (LSTM) cells \cite{hochreiter1997} and 
the attention mechanism  \cite{bahdanau2014neural}. %,luong-pham-manning:2015:EMNLP}. 
This evolution has several implications.
On one side, NMT represents a simplification with respect to previous paradigms.  From a management point of view, similar to PBMT, it allows for a more efficient use of human and data resources with respect to rule-based MT. From the architectural point of view,  a large recurrent network trained for end-to-end translation is considerably simpler than traditional MT systems that integrate multiple components and processing steps. 
On the other side, the NMT process is less transparent than previous paradigms. Indeed, it represents a further step in the evolution from rule-based approaches that explicitly manipulate knowledge, to the statistical/data-driven framework, still comprehensible in its inner workings, to a sub-symbolic framework in which the translation process is totally opaque to the analysis.

What do we know about the strengths of NMT and the weaknesses of PBMT? What are the linguistic phenomena that deep learning translation models can handle with such greater effectiveness?
%In this paper we address these questions, trying to understand what makes one approach better than the other. Going beyond scarcely informative BLEU scores, 
To answer these questions and go beyond poorly informative BLEU scores,
we perform the very first comparative analysis of the two paradigms in order to shed light on the factors that differentiate them and determine their large quality differences.
%of the type of errors that differentiate the two paradigms and determine their large performance differences.

We build on evaluation data available for the IWSLT 2015 MT English-German task, and compare the results of the first four top-ranked participants. 
%
%Our choice to focus on one single task addressing only one language pair, instead of aiming at a more large-scale and representative evaluation, is motivated by the distinguishing features of the chosen task, which provide a particularly favourable evaluation setting for an in-depth and fine-grained analysis: 
%Instead of aiming at a more large-scale and representative evaluation,
We choose to focus on one language pair and one task because of the following advantages:
\textit{(i)} three state-of-the art PBMT systems compared against the NMT system on the same data and in the very same period (that of the evaluation campaign); \textit{(ii)} a challenging language pair in terms of morphology and word order differences; %\textit{(iii)} human evaluation based on MT outputs' post-editing done by professional translators, 
\textit{(iii)} availability of MT outputs' post-editing done by professional translators, 
which is very costly and thus rarely available. In general, post-edits have the advantage of allowing for informative and detailed analyses since they directly point to translation errors. In this specific framework, the high quality data  created by professional translators guarantees reliable evaluations.
For all these reasons we present our study as a solid contribution to the better understanding of this new paradigm shift in MT.

After reviewing previous work  (Section \ref{previous}),  we introduce 
the analyzed data and  the systems that produced them (Section \ref{exp-setting}). We then present three increasingly fine levels of MT quality analysis.  We first investigate how MT systems' quality varies with specific characteristics of the input,  \textit{i.e.} sentence length and type of content of each talk (Section \ref{overall-quality}).
%%analyse/investigate/explore
Then, we focus on differences among MT systems with respect to morphology, lexical, and word order errors (Section \ref{errors}).
%%to understand which types of errors characterize NMT with respect to SMT (Section \ref{errors}).
%%concentrate/focus/measure/
Finally, based on the finding that word reordering is the strongest aspect of NMT compared to the other systems, we carry out a fine-grained analysis of word order errors 
%, which we deem particularly relevant also given the language pair addressed, that is characterized by %important word order differences 
(Section \ref{fine_reordering}).

\section{Previous Work}
\label{previous}

%AGGIUNGERE REFERENCE
%described results applying neural MT reranking to a baseline syntax-based machine translation system in 4 languages. In particular, we performed an in-depth analysis of what kinds of translation errors were fixed by neural MT reranking. Based on this analysis, we found that the majority of the gains were related to improvements in the accuracy of transfer of correct grammatical structure to the target sentence, with the most prominent gains being related to errors regarding reordering of phrases, insertion/deletion of copulas, coordinate structures, and verb agreement.

To date, NMT systems have only been evaluated by BLEU in single-reference setups
\cite{bahdanau2014neural,Sutskever14,luong-pham-manning:2015:EMNLP,Jean:15,Gulcehre:15}.
Additionally, the Montreal NMT system submitted to WMT 2015 \cite{WMT-15-montreal} was part of a manual evaluation experiment where a large number of non-professional annotators were asked to rank the outputs of multiple MT systems
%, while being shown the source and reference translation 
\cite{WMT-15}.
Results for the Montreal system were very positive -- ranked first in English-German, third in German-English, English-Czech and Czech-English -- which confirmed and strengthened the BLEU results published so far.
Unfortunately neither BLEU nor manual ranking judgements tell us which translation aspects are better modeled by different MT frameworks.
% NMT than others, or in other words, what are the strengths and weaknesses of NMT. % with respect to SMT.
To this end, %we set out to conduct 
a detailed and systematic error analysis of NMT \textit{vs.} PBMT output is required.
% is needed, which is the goal of this paper.

Translation error analysis, as a way to identify systems' weaknesses and define priorities for their improvement, has received a fair amount of attention in the MT community. 
%There is bulk of work that both with automatic and manual methods has been trying to profile  the behaviour of MT systems with respect to various typologies of errors.
%
In this work we opt for the \textit{automatic} detection and classification of translation errors 
based on \textit{manual} post-edits of the MT output. 
We believe this choice provides an optimal trade-off between 
fully manual error analysis \cite{farrus:10,popovic13,daems:2014,Federico:14,neubig-morishita-nakamura:2015:WAT}, which is very costly and complex, 
and fully automatic error analysis \cite{popovic:11,irvine:13}, % based on reference translations, 
which is noisy and biased towards one or few arbitrary reference translations. % choices of one or few translators.
%difficult to interpret.

%\cite{Bojar:11}: proposes a method for interpreting blind post-editing data at a finer level (and compare the results with explicit marking of errors). Systems participating in WMT9 Eng-Czech were analysed exploiting their corresponding blind post-editing. A simple methodology based on editing operations was followed, where editing operations are mapped into errors.\\

%TODO: cite Koponen 12,

Existing tools for translation error detection are either based on Word Error Rate (WER) and Position-independent word Error Rate (PER) \cite{popovic:11:hjerson} or on 
%a user-provided 
output-reference alignment \cite{zeman:11}.
%computed, for instance, using bilingual GIZA++ and pivoting through the source sentence.
%
Regarding error classification, Hjerson \cite{popovic:11:hjerson} detects five main types of word-level errors as defined in \cite{Vilar:06a}: morphological, reordering, missing words, extra words, and lexical choice errors.
We follow a similar but simpler error classification (morphological, lexical, and word order errors), but detect the errors differently using TER as this is the most natural choice in our evaluation framework based on post-edits (see also Section~\ref{methodology}).
%
% a metric typically used to judge speech recognition output and not suitable to account for reordering errors.
%Hjerson \cite{popovic:11:hjerson} is an open source tool for automatic classification of MT output, which covers the main word level error categories defined in \cite{Vilar:06a}: morphological, reordering, missing words, extra words, and lexical choice errors. It implements a method based on the standard WER combined with the precision and recall based error rates. Can work both on single references and multiple references.\\
%Addicter \cite{zeman:11} is a similar tool that further allows the user to choose among different output-reference alignment techniques or to use an external alignment tool such as GIZA++. Errors are then detected from the selected alignment.
%\cite{zeman:11} ADDICTER: open-source tool that uses a method based explicilty on aligning the hypothesis and reference translations to devise the various error types from there.\\
%
\newcite{irvine:13} propose another word-level error analysis technique specifically focused on lexical choice and aimed at understanding the effects of domain differences on MT.
Their error classification is strictly related to model coverage and insensitive to word order differences.
The technique requires access to the system's phrase table and is thus not %directly 
applicable to NMT, which does not rely on a fixed inventory of translation units extracted from the parallel data.

Previous error analyses based on manually post-edited translations were presented in \cite{Bojar:11,Koponen:12,popovic13}. 
We are the first to conduct this kind of study on the output of a neural MT system.

\section{Experimental Setting}
\label{exp-setting}

We perform a number of analyses %aimed at understanding the behaviour of NMT and comparing it to the other approaches with respect to factors that can influence translation quality. These analyses build 
on data and results of the IWSLT 2015 MT \textit{En-De} task, which consists in translating
% of TED talks, \textit{i.e.} public speeches covering different topics. The MT \textit{En-De} task consists in translating 
manual transcripts of English TED talks into German.
Evaluation data are publicly available through the {\scshape WIT$^{\bf 3}$} repository \cite{Cettolo:2012}.\footnote{wit3.fbk.eu}

\subsection{Task Data}
\label{task-data}

TED Talks\footnote{http://www.ted.com/}  
are a collection of rather short speeches (max 18 minutes each, roughly equivalent to 2,500 words) covering a wide variety of topics. 
All talks have captions, which are translated into many languages by volunteers worldwide.
Besides representing a popular benchmark for spoken language technology, TED Talks embed interesting research challenges. 
%From the point of view of MT, 
Translating TED Talks implies dealing with spoken rather than written language, which is hence expected to be structurally less complex, formal and fluent \cite{ruiz:2014}. Moreover, as human translations of the talks are required to follow the structure and rhythm of the English captions, a lower amount of rephrasing and reordering is expected than in the translation of written documents.

As regards the English-German language pair, the two languages are interesting since, while belonging to the same language family, they have marked differences in levels of inflection, morphological variation, and word order, especially long-range reordering of verbs.

\subsection{Evaluation Data}
\label{data}

Five systems participated in the MT \textit{En-De} task and were manually evaluated on a representative subset of the official 2015 test set. The Human Evaluation (HE) set includes %around the initial 56\% approximately 
the first half of each of the 12 test talks, for a total of 600 sentences and around 10K words.
%
%The output of the five systems on the HE set was assigned to 
Five professional translators were asked to post-edit the MT output by applying the minimal edits required to transform it into a fluent sentence with the same meaning as the source sentence.
Data were prepared so that all translators equally post-edited the five MT outputs, \textit{i.e.} 120 sentences for each evaluated system.

The  resulting evaluation data consist of five new reference translations for each of the sentences in the HE set. Each one of these references represents the \textit{targeted translation} of the system output from which it was derived, but the other four \textit{additional translations} can also be used to evaluate each MT system.
We will see in the next sections how we exploited the available post-edits in the more suitable way depending on the kind of analysis carried out.

\subsection{MT Systems}
\label{systems}

\begin{table}[tb]
    \setlength{\tabcolsep}{2pt}
    \footnotesize
    \begin{tabular}{|c| l | l |}
    \hline
      System       &     \multicolumn{1}{c|}{Approach}       & Data \\
    \hline
      \uedin & Combination: Phrase+Syntax-based  & \footnotesize 175M/   \\
             \footnotesize (Huck and       & \footnotesize GHKM string-to-tree; hierarchical +  & \footnotesize 3.1B \\
              \footnotesize Birch, 2015)    & \footnotesize sparse lexicalized reordering models &  \\
    \hline
      \hdu   & Hierarchical  Phrase-based & 166M/      \\
            \footnotesize (Jehl et al., & \footnotesize  source pre-ordering (dependency tree & 854M \\
            \footnotesize 2015)  & \footnotesize -based); re-scoring with neural LM  & \\
    \hline
    \kit      & Standard Phrase-based  & 117M/   \\
            \footnotesize (Ha et al.,     & \footnotesize source pre-ordering (POS- and tree- & 2.4B  \\
             \footnotesize 2015)     & \footnotesize based); re-scoring with neural LMs & \\
    \hline
          \su   & Recurrent neural network (LSTM) & 120M/       \\
            \footnotesize (Luong \& Man-    & \footnotesize attention-based; source reversing;   &  --     \\
            \footnotesize ning,  2015)     & \footnotesize rare words handling   &       \\
    \hline
    \end{tabular}
    \caption{MT systems' overview. Data column: size of parallel/monolingual training data for each system in terms of English and German tokens.}
    \label{tab:systems}
\end{table}

Our analysis focuses on the first four top-ranking systems, which include NMT \cite{luong2015stanford} and three different phrase-based approaches: standard phrase-based \cite{iwslt-kit-mt:15},  hierarchical~\cite{iwslt-hdu:15}
and a combination of phrase-based and syntax-based
\cite{iwslt-uedin:15}. 
Table \ref{tab:systems} presents an overview of each system, as well as figures about the training data used.\footnote{Detailed information about training data was kindly made available by participating teams.} 
%CAMERA-READY

The phrase+syntax-based (\uedin) system combines the outputs of a string-to-tree decoder, trained with the GHKM algorithm, with those of two standard phrase-based systems featuring, among others, adapted phrase tables and language models enriched with morphological information, hierarchical lexicalized reordering models %models, lexicalized sparse reordering features, 
and different variations of the operational sequence model.  

The hierarchical phrase-based MT (\hdu) system leverages thousands of lexicalised features, 
data-driven source pre-ordering (dependency tree-based),
word-based and class-based language models, and n-best re-scoring models 
based on syntactic and neural language models. 

The standard phrase-based MT (\kit) system features an adapted phrase-table combining in-domain and out-domain data, discriminative word lexicon models, multiple language models (word-, POS- and class-based), data-driven source pre-ordering (POS- and constituency syntax-based), n-best re-scoring models based on neural lexicons and neural language models. 

Finally, the neural MT (\su) system is an ensemble of 8 long short-term memory (LSTM) networks of 4 layers featuring 1,000-dimension word embeddings, attention mechanism, source reversing, 50K source and target vocabularies, and out-of-vocabulary word handling. Training with TED data was performed on top of models trained with large out-domain parallel data.  

With respect to the use of training data, it is worth noticing that 
{\su } is the only system not employing monolingual data in addition to parallel data. %,  did not leverage any monolingual target language data to train a language model. 
Moreover, {\su } and {\kit } were trained with smaller amounts of parallel data with respect to 
{\uedin } and {\hdu } (see Table~1).

%\subsection{Evaluation Metrics}
\subsection{\AB{Translation Edit Rate Measures}}
\label{methodology}

The \textit{Translation Edit Rate} (TER)~\cite{Snover:06}
\COMMENT{
results. TER~\cite{Snover:06} measures the amount of editing that a human would have to perform to change an automatic translation so that it exactly matches a given reference translation. TER distinguishes among four different types of edit operations: {\it deletion}, {\it insertion}, {\it substitution}, {\it shift}. An important characteristic of {\it shift} operations is that movements of blocks are allowed and counted as one edit.
}
naturally fits our evaluation framework, %based on post-edits since, in this scenario, 
where it traces the edits done by post-editors. Also, TER {\it shift} operations are reliable indicators of reordering errors, in which we are particularly interested.
We exploit the available post-edits in two different ways: %, thus relying on different measures, namely 
\textit{(i)} for \textit{Human-targeted TER} (HTER) we compute TER between the machine translation and its manually post-edited version (targeted reference), \textit{(ii)} for \textit{Multi-reference TER} (mTER), we compute TER against the closest translation among all available post-edits (\textit{i.e.} targeted and additional references) for each sentence.

Throughout sections~\ref{overall-quality} and \ref{errors}, we mark a score achieved by NMT with the symbol~* if this is better than the score of its best competitor at statistical significance level $0.01$.
Significance tests for HTER and mTER are computed by bootstrap re-sampling, while differences among proportions are assessed via one-tailed \mbox{z-score} tests. 

\section{Overall Translation Quality}
\label{overall-quality}

Table~\ref{tab:hter} presents overall system results according to HTER and mTER, as well
as BLEU computed against the original TED Talks reference translation.
We can see that NMT clearly outperforms all other approaches both in terms of BLEU and TER scores. Focusing on mTER results, the gain obtained by {\su} over the second best system (\uedin) amounts to 26\%.
It is also worth noticing that mTER is considerably lower than HTER for each system.
This reduction shows that exploiting all the available post-edits as references for TER is a viable way to control and overcome post-editors  variability, thus ensuring a more reliable and informative evaluation about the real overall performance of MT systems. 
For this reason, the two following analyses rely on mTER. % results calculated on all the available post-edits.
%\footnote{As described in Section \ref{data}, five post-edits were created for the official IWSLT human evaluation because five systems participated in the task.} 
In particular, we investigate how specific characteristics of input documents affect the system's overall translation quality, focusing on \textit{(i)} sentence length and \textit{(ii)} the different talks composing the dataset.

\begin{table}[t]
    \small\centering
    \begin{tabular}{l|lll}
    \hline
    system & BLEU & HTER & mTER \\
    \hline
    %\pjait & 35.7 & 28.2 \\
    \uedin   & 25.3  & 28.0 & 21.8 \\
    \hdu   & 24.6  & 29.9 & 23.4 \\
    \kit   & 25.8  & 29.0 & 22.7 \\
    \su    & 31.1$^*$  & 21.1$^*$ & 16.2$^*$ \\
    \hline
    \end{tabular}
    \caption{Overall results on the HE Set: BLEU, computed against the original reference translation, and TER, computed with respect to the targeted post-edit (HTER) and multiple post-edits (mTER).} 
    \label{tab:hter}
\end{table}

\subsection{Translation quality by sentence length}

\begin{figure}[t]
\centering
%\hspace{-6mm}
\includegraphics[width=\columnwidth]{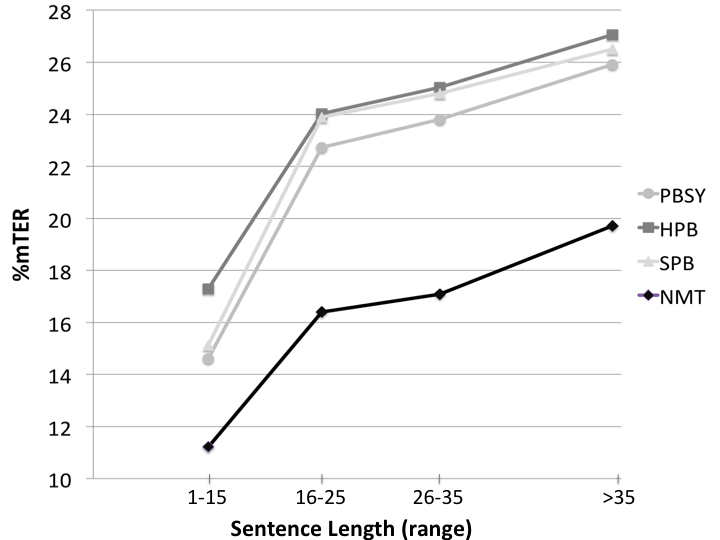}
%{scoresPerLen_4bins.jpg}
%\vspace{-12mm}
\caption{mTER scores on bins of sentences of different length. Points represent the average mTER of the MT outputs for the sentences in each given bin. }
\label{fig:mhterPerLen}
%\vspace{-5mm}
\end{figure}

Long sentences are known to be difficult to translate by the NMT approach. 
Following previous work \cite{cho2014properties,pougetabadie-EtAl:2014:SSST-8,bahdanau2014neural,luong-pham-manning:2015:EMNLP}, we investigate how sentence length affects overall translation quality.
%
%Sentence length represents an issue for the NMT approach, which typically suffers when long sentences are to be translated, in contrast to SMT systems which are expected to  be more  robust~\cite{cho2014properties,pougetabadie-EtAl:2014:SSST-8,bahdanau2014neural}.
%translation  performance  does   drop  for  sentences of  lengths  greater  than 
%
%Input sentence length has been considered by a number of studies aiming to improve NMT's ability to handle long sentences~\cite{cho2014properties,pougetabadie-EtAl:2014:SSST-8,bahdanau2014neural,luong-pham-manning:2015:EMNLP}.
%preventing NMT systems from underperforming with long sentences.
%
Figure~\ref{fig:mhterPerLen} plots mTER scores against source sentence length. 
{\su} clearly outperforms every PBMT system in any length bin, with statistically significant differences. As a general tendency, the performance of all approaches worsens as sentence length increases. However, for sentences longer than 35 words we see that {\su} quality degrades more markedly than in PBMT systems. 
Considering the percentage decrease with respect to the preceding length bin (26-35), we see that the $\%\Delta$ for {\su}  (-15.4) is much larger than the average $\%\Delta$ for the three PBMT systems  (-7.9).
%
%The percentage decrease between the $>$35 bin and the 26-35 bin, and we see that the $\%\Delta$ for {\su} (-15.4) is much higher than the average $\%\Delta$ for the three PBMT systems.
%
%The percentage decrease with respect to the preceding length bin (26-35) amounts to 15.4 for {\su}, which is much higher than the average $\%\Delta$ for the three PBMT systems
%
Hence, this still seems an issue to be addressed for further improving {\su}. 

\subsection{Translation quality by talk}

As we saw in Section \ref{task-data}, the TED dataset is very heterogeneous since it consists of talks covering different topics and given by speakers with different styles. %, with different lexica, and who construct sentences in different ways. 
%
%We investigate whether specific characteristics of the talks affect overall translation performance. 
%
It is therefore interesting to evaluate translation quality also at the talk level.

\COMMENT{
On all the twelve talks included in the HE set, the NMT system outperforms the PBMT systems in a statistically significant way. Moreover, we find a moderate Pearson correlation (R=$0.7332$) between the type-token ratio\footnote{The type-token-ratio of a text is calculated dividing the number of word types (vocabulary) by the total number of word tokens (occurrences).} and the mTER gains of NMT over its closest competitor in each talk. 
This result suggests that NMT is able to cope with lexical diversity better than any other considered approach.
}

\begin{figure}[t]
%\centering
%\hspace{-6mm}
%\includegraphics[width=\columnwidth]{scoresPerTalk.pdf}
%\vspace*{-5mm}\hspace*{-0.8cm}
\includegraphics[width=\columnwidth]{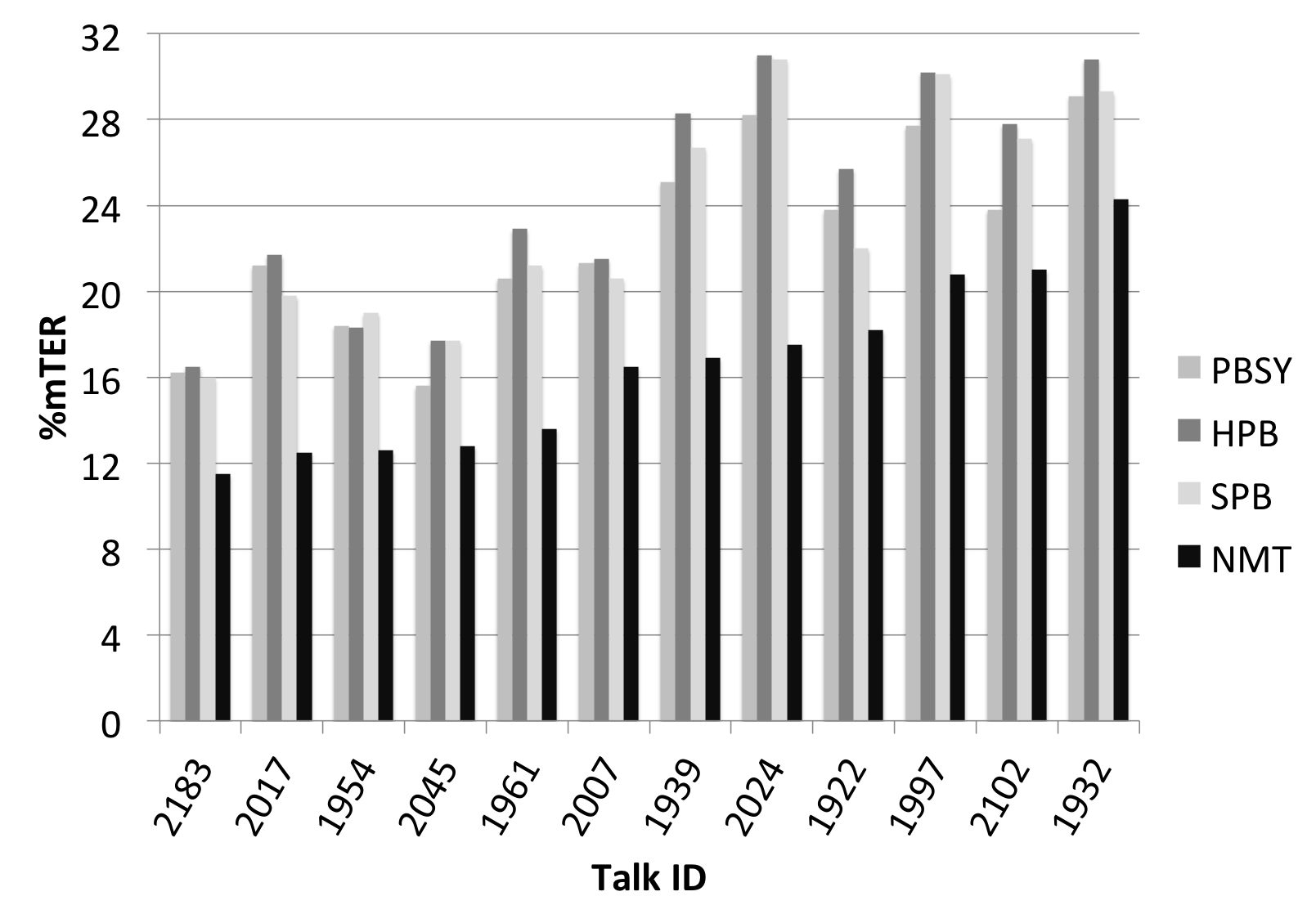}
%\vspace{-12mm}
\caption{mTER scores per talk, sorted in ascending order of NMT scores.}
\label{fig:mhterPerTalk}
\end{figure}

Figure~\ref{fig:mhterPerTalk} plots the mTER scores for each of the twelve talks included in the HE set, sorted in ascending order of {\su } scores. In all talks, the NMT system outperforms the PBMT systems in a statistically significant way.

%It can be noted that in most cases the NMT system clearly outperforms all the PBMT systems, 
%%(e.g. in talks 2024, 2017, 1961)
%while in few cases the gap is smaller (\textit{e.g.} 2102, 2045, 1922), even though still statistically significant. 

We analysed different factors which could impact translation quality in order to understand if they correlate with such performance differences.
We studied three features which are typically considered as indicators of complexity (see \cite{franccois-fairon:2012:EMNLP-CoNLL} for an overview), namely \textit{(i)} the length of the talk, \textit{(ii)} its average sentence length, and \textit{(iii)} the type-token ratio\footnote{The type-token-ratio of a text is calculated dividing the number of word types (vocabulary) by the total number of word tokens (occurrences).} (TTR) which -- measuring lexical diversity -- reflects the size of a speaker's vocabulary and the variety of subject matter in a text.  
      
For the first two features we did not find any correlation; on the contrary, we found a moderate Pearson correlation (R=$0.7332$) between TTR and the mTER gains of NMT over its closest competitor in each talk. 
This result suggests that NMT is able to cope with lexical diversity better than any other considered approach.

\COMMENT{
\begin{table}[ht!]
    \setlength{\tabcolsep}{6pt}
    \small\centering
    \begin{tabular}{l|cccc}
\normalsize talkId & \small \uedin & \small \hdu & \small\kit & \small \su \\
\hline
1922	 & 23.8 & 25.7 & 22.0 & 18.2 \\
1932	 & 29.1 & 30.8 & 29.3 & 24.3 \\
1939	 & 25.1 & 28.3 & 26.7 & 16.9 \\
1954	 & 18.4 & 18.3 & 19.0 & 12.6 \\
1961	 & 20.6 & 22.9 & 21.2 & 13.6 \\
1997	 & 27.7 & 30.2 & 30.1 & 20.8 \\
2007	 & 21.3 & 21.5 & 20.6 & 16.5 \\
2017	 & 21.2 & 21.7 & 19.8 & 12.5 \\
2024	 & 28.2 & 31.0 & 30.8 & 17.5 \\
2045	 & 15.6 & 17.7 & 17.7 & 12.8 \\
2102	 & 23.8 & 27.8 & 27.1 & 21.0 \\
2183	 & 16.2 & 16.5 & 16.0 & 11.5 \\
    \hline
    \end{tabular}
    \caption{mTER scores on each talk.}
    \label{tab:mhterTalks}
\end{table}
}

\section{Analysis of Translation Errors}
\label{errors}
%After comparing systems' overall performance according to source sentence length and talk, we go deeper into the evaluation, aiming 
We now turn to analyze which types of linguistic errors characterize NMT \textit{vs.} PBMT.
In the literature, various error taxonomies covering different levels of granularity have been developed \cite{Flanagan:94,Vilar:06a,farrus:10,Stymne:12,Lommel:14}.
We focus on three error categories, namely \textit{(i)} morphology errors, \textit{(ii)} lexical errors, and \textit{(iii)} word order errors. 
As for lexical errors, a number of existing taxonomies further distinguish among translation errors due to missing words, extra words, or incorrect lexical choice. However, given the proven difficulty of disambiguating between these three subclasses \cite{popovic:11,fishel:2012}, we prefer to rely on a more coarse-grained linguistic error classification where lexical errors include all of them \cite{farrus:10}.

For error analysis we rely on HTER results under the assumption that, since the targeted translation is generated by post-editing the given MT output, this method is particularly informative to spot MT errors. 
We are aware that translator subjectivity is still an issue (see Section \ref{overall-quality}), however in this more fine-grained analysis we prefer to focus on what a human implicitly annotated as a translation error. %, whereas the most realistic overall performance of the system -- as given by mTER -- is less relevant.  CAMERA-READY
This particularly holds in our specific evaluation framework, where the goal is not to measure the absolute number of errors made by each system, but to compare systems with each other. Moreover, the post-edits collected for each MT output within IWSLT allow for a fair and reliable comparison since systems were equally post-edited by all translators (see Section \ref{data}), making all analyses uniformly affected by such variability.

\subsection{Morphology errors}

%A morphology error occurs when a generated word form is wrong but its corresponding base form (lemma) is correct. Thus, the ability of a system to deal with morphology can be indirectly assessed by comparing the  score of any MT metric computed on the surface form of the words (i.e. the morphologically inflected words) with the score on their corresponding lemmas (i.e. the base form without morphological inflection): the closer the two scores, the more accurate the morphology. In the best case were there are no morphology errors, no additional matches are counted on lemmas with respect to word forms and the two scores are equal.

A morphology error occurs when a generated word form is wrong but its corresponding base form (lemma) is correct. Thus, we assess the ability of systems to deal with morphology by comparing the HTER score computed on the surface forms (\textit{i.e.} morphologically inflected words) with the HTER score obtained on the corresponding lemmas. % (\textit{i.e.} base forms without morphological inflection). 
The additional matches counted on lemmas with respect to word forms indicate morphology errors. Thus, the closer the two HTER scores, the more accurate the system in handling morphology.

To carry out this analysis, the lemmatized (and POS tagged) version of both MT outputs and corresponding post-edits was produced with the German parser ParZu %\footnote{https://github.com/rsennrich/parzu} 
\cite{parzu-de:13}. 
Then, the HTER-based evaluation was slightly adapted in order to be better suited to an accurate detection of morphology errors.
First, punctuation was removed since  -- not being subject to morphological inflection -- it  could smooth the results.
Second, {\it shift} errors were not considered.
A word form or a lemma that matches a corresponding word or lemma in the post-edit, but is in the wrong position with respect to it, is counted as a \textit{shift} error in TER. Instead -- when focusing on morphology -- exact matches are not errors, regardless their position in the text.\footnote{Note that the TER score calculated by setting to 0 the cost of shifts approximates the Position-independent Error Rate~\cite{Tillmann97}.} 
%\textcolor{red}{@Mauro: hai messo a 0 gli shift anche quando hai fatto il TER per le word form, vero? Mauro: s\`i}

\begin{table}[t]
    \small\centering
    \begin{tabular}{l|lll}
    \hline
    \multirow{2}{*}{system}    & \multicolumn{2}{c}{HTER{noShft}} & \\
           & word & lemma &  \%$\Delta$ \\
    \hline
    
    \uedin & 27.1 & 22.5 & -16.9 \\
    \hdu   & 28.7 & 23.5 & -18.4 \\
    \kit   & 28.3 & 23.2 & -18.0 \\
    \su    & 21.7$^*$ & 18.7$^*$ & -13.7 \\
    \hline
    \end{tabular}
    \caption{HTER ignoring shift operations %(HTERnoShft) 
    computed on words and corresponding lemmas, and their \% difference.}
    \label{tab:wrdVSlmm-IDS}
\vspace{-2mm}
\end{table}

%Ecco i conteggi relativi alla
%Table 4
%		word				lemma
%PBSY  	(2954-316)/9730	(2568-377)/9730
%HIER  	(3147-379)/9629	(2715-456)/9629
%PBMT  	(3055-343)/9596	(2624-399)/9596
%NMT   	(2225-160)/9534	(1966-184)/9534

Table~\ref{tab:wrdVSlmm-IDS} presents  HTER scores on word forms and  lemmas, as well as their percentage difference which gives an indication of morphology errors.
We can see that {\su } generates translations which are morphologically more correct than the other systems. In particular, 
the $\%\Delta$ for {\su } (-13.7) is lower  than that of the second best system (\uedin, -16.9) by 3.2\% absolute points, leading to a percentage gain of around 19\%. We can thus say that {\su } makes at least 19\% less morphology errors than any other PBMT system.
%We conclude that also under this aspect the neural approach succeeds, but not as much as for  reordering words.

\subsection{Lexical errors}

Another important feature of MT systems is their ability to choose lexically appropriate words. In order to compare systems under this aspect, we consider HTER results at the lemma level as a way to abstract from morphology errors and focus only on actual lexical choice problems. 
The evaluation on the lemmatised version of the data performed to identify morphology errors fits this purpose, since its driving assumptions (\textit{i.e.} punctuation can be excluded and lemmas in the wrong order are not errors) hold for lexical errors too. 

%Hence, we can look again at
The lemma column of Table~\ref{tab:wrdVSlmm-IDS} shows that {\su } outperforms the other systems. More precisely, the {\su } score (18.7) is better than the second best (\uedin, 22.5)  by 3.8\% absolute points. This corresponds to a relative gain of about 17\%, meaning that {\su } makes  at least 17\% less lexical errors than any PBMT system. 
Similarly to what observed for morphology errors, this can be considered a remarkable improvement over the state of the art.
%, although not outstanding like that observed for word reordering.

\subsection{Word order errors}
\label{sect:shifts}

%\AB{Arianna: anche in questa sezione ignoriamo la punteggiatura, giusto? MAURO: NO, QUI LA PUNTEGGIATURA RIENTRA NEI CONTI.}

\begin{table}[t]
    \small\centering
    \begin{tabular}{l|ccc|c}
    \hline
    system  & \#words & \#shifts & \%shifts & KRS \\
        \hline
    \uedin &  11,517 & 354 & 3.1 & 84.6 \\
    \hdu   &  11,417 & 415 & 3.6  & 84.3 \\
    \kit     &  11,420 & 398 & 3.5  & 84.5 \\
    \su    &  11,284 & 173 & \ 1.5$^*$ & \ 88.3$^*$ \\
    \hline
    \end{tabular}
    \caption{Word reordering evaluation in terms of shift operations in HTER calculation and of KRS. For each system, the number of generated words, the number of shift errors and their corresponding percentages are reported.} % Statistical significance between {\su} and {\uedin} is assessed with a one-tailed proportion test. }
    \label{tab:reorderingHTER}
\end{table}

%Conteggi relativi alla Table
%        	  #Shft  #Err  #wrds
%PBSY	  354    3224  11517
%HIER	  415    3417  11417
%PBMT	  398    3310  11420
%NMT		  173    2380  11284

% KRS statistical significance (computed by approximate randomization with the sigf tool):
% \su vs ALL: p-value < 0.01
% \uedin vs \hdu: p-value=0.59
% \uedin vs \kit: p-value=0.87
% \hdu vs \kit: p-value=0.74

To analyse reordering errors, we start by focusing on {\it shift} operations identified by the HTER metrics. The first three columns of Table~\ref{tab:reorderingHTER}   show, respectively: \textit{(i)} the number of words generated by each system \textit{(ii)} the number of shifts required to align each system output to the corresponding post-edit; and \textit{(iii)} the corresponding percentage of shift errors. Notice that the \textit{shift} error percentages are incorporated in the HTER scores reported in Table~2. 
We can see in Table~\ref{tab:reorderingHTER} that \textit{shift} errors in {\su } translations are definitely less than in the other systems. The error reduction of {\su } with respect to the second best system (\uedin) is about 50\% (173 \textit{vs.} 354).

It should be recalled that these numbers only refer to \textit{shifts} detected by HTER, that is (groups of) words of the MT output and corresponding post-edit that are identical but occurring in different positions.
Words that had to be moved and modified at the same time (for instance replaced by a synonym or a morphological variant) are not counted in HTER \textit{shift} figures, but are detected as \textit{substitution}, \textit{insertion} or \textit{deletion} operations.
To ensure that our reordering evaluation is not biased towards the alignment between the MT output and the post-edit performed by HTER, we run an additional assessment using KRS -- Kendall Reordering Score \cite{Birch:10} -- which measures the similarity between the source-reference reorderings and the source-MT output reorderings.\footnote{To compute the word alignments required by KRS, we used the FastAlign tool \cite{fastalign:13}.}
Being based on bilingual word alignment via the source sentence, KRS detects reordering errors also when post-edit and MT words are not identical. Also unlike TER, KRS is sensitive to the \textit{distance} between the position of a word in the MT output and that in the reference. 

Looking at the last column of Table~\ref{tab:reorderingHTER}, we can say that our observations on HTER are confirmed by the KRS results: the reorderings performed by NMT are much more accurate than those performed by any PBMT system.% 
\footnote{To put our results into perspective, note that
\newcite{Birch:11:phdthesis} reports a difference of 5 KRS points between the translations of a PBMT system and those produced by four \textit{human} translators tested against each other, in a Chinese-English experiment.}
%We compare the reorderings seen in the output of the translation models to the human translated references and to each other.
%Note that a difference of 4 KRS points indicates a remarkable difference in the quality of word reordering. \textcolor{red}{@Arianna: any reference here?}
Moreover, according to the approximate randomization test, KRS differences are statistically significant between \su\ and all other systems, but not among the three PBMT systems.

Given the concordant results of our two quantitative analyses, we conclude that one of the major strengths of the {\su } approach is its ability to place German words in the right position even when this requires considerable reordering. This outcome calls for a deeper investigation, which is carried out in the following section.

\section{Fine-grained Word Order Error Analysis}
\label{fine_reordering}

%In Section~\ref{sect:shifts} 
We have observed that word reordering is a very strong aspect of NMT compared to PBMT, according to both HTER and KRS.
To better understand this finding, we investigate whether reordering errors concentrate on specific linguistic constructions across our systems.
%
% CAMERA-READY: new-paragpraph To this end 
%We produce POS tagging and dependency parsing of the post-edits using ParZu \cite{parzu-de:13} Then, we use this annotation to 
Using the POS tagging and dependency parsing of the post-edits produced by ParZu, % \cite{parzu-de:13},
we classify the \textit{shift} operations detected by HTER 
and count how often a word with a given POS label was misplaced by each of the systems (alone or as part of a shifted block).
For each word class, we also compute the percentage order error reduction of NMT with respect to the PBMT system that has highest reordering accuracy overall, that is {\uedin}.
Results are presented in Table~\ref{tab:shiftClasses}, ranked by NMT-vs-\uedin\ gain.
Punctuation is omitted as well as word classes that were shifted less than 10 times by all systems.
Examples of salient word order error types are presented in Table~\ref{table:examples}.

\setlength{\tabcolsep}{1.8pt}
\begin{table}[t]
\centering\small
\begin{tabular}{@{\ }c|c|c@{\ \ }ccc@{\ }}
\hline
\multirow{2}{*}{Class} & NMT- & \multirow{2}{*}{\su}  & \multirow{2}{*}{\uedin} & \multirow{2}{*}{\hdu} & \multirow{2}{*}{\kit} \\
  & vs-\uedin & & & & \\
%  \%$\Delta$(SYNT,NMT)
\hline
%### Shift POStags:											
%[PAT]	&	sta-vs-UEDIN	&	sta	&	UED	&	hdu	&	kit	\\
\sc v	&	-70\%	&	35	&	116	&	133	&	155	\\
%\depgap\depfont (aux:\textsc{v})	& \depfont -87\%	& \depfont 	3	&	\depfont 23	&	\depfont 17	&	\depfont 18	\\
%\depgap\depfont (neb:\textsc{v})	&\depfont 	-83\%	&\depfont	2	&\depfont	12	&\depfont	7	&\depfont	19	\\
%\depgap\depfont (objc:\textsc{v})	&\depfont	-79\%	&\depfont	3	&\depfont	14	&\depfont	21	&\depfont	24	\\
% MERGED root and mroot:
%root:\textsc{v})	&	-68\%	&	6	&	19	&	28	&	27	\\
%mroot:\textsc{v}	&	-36\%	&	7	&	11	&	26	&	20	\\
%\depgap\depfont (root:\textsc{v})	&\depfont	-57\%	&\depfont	13	&\depfont	30	&\depfont	54	&\depfont	47	\\
%\depgap\depfont  (cj:\textsc{v})	&\depfont	-59\%	&\depfont	7	&\depfont	17	&\depfont	21	&\depfont	22	\\
\sc pro	&	-57\%	&	22	&	51	&	53	&	62	\\
\sc ptkzu	&	-54\%	&	6	&	13	&	4	&	11	\\
\sc adv	&	-50\%	&	14	&	28	&	44	&	36	\\
\sc n	&	-47\%	&	37	&	70	&	99	&	56	\\
%\depgap\depfont obja:\textsc{n}	&\depfont	-65\%	&\depfont	6	&\depfont	17	&\depfont	28	&\depfont	12	\\
%\depgap\depfont pn:\textsc{n}	&\depfont	-36\%	&\depfont	16	&\depfont	25	&\depfont	33	&\depfont	19	\\
%\depgap\depfont subj:\textsc{n}	&\depfont	-33\%	&\depfont	6	&\depfont	9	&\depfont	10	&\depfont	7	\\
%$,	&	-39\%	&	34	&	56	&	49	&	58	\\
\sc kon	&	-33\%	&	6	&	9	&	8	&	12	\\
\sc prep	&	-18\%	&	18	&	22	&	27	&	28	\\
\sc ptkneg	&	-17\%	&	10	&	12	&	10	&	7	\\
\sc art	&	-4\%	&	26	&	27	&	38	&	35	\\
\hline
%### Shift DEPrels:											
%[PAT]	&	sta-vs-UEDIN	&	sta	&	UED	&	hdu	&	kit	\\
aux:\textsc{v}	&	-87\%	&	3	&	23	&	17	&	18	\\
neb:\textsc{v}	&	-83\%	&	2	&	12	&	7	&	19	\\
objc:\textsc{v}	&	-79\%	&	3	&	14	&	21	&	24	\\
subj:\textsc{pro}	&	-70\%	&	12	&	40	&	34	&	46	\\
root:\textsc{v}	&	-68\%	&	6	&	19	&	28	&	27	\\
adv:\textsc{adv}	&	-67\%	&	8	&	24	&	33	&	28	\\
obja:\textsc{n}	&	-65\%	&	6	&	17	&	28	&	12	\\
cj:\textsc{v}	&	-59\%	&	7	&	17	&	21	&	22	\\
part:\textsc{ptkzu}	&	-54\%	&	6	&	13	&	4	&	11	\\
%punc:$,	&	-39\%	&	34	&	56	&	49	&	58	\\
obja:\textsc{pro}	&	-38\%	&	5	&	8	&	14	&	7	\\
mroot:\textsc{v}	&	-36\%	&	7	&	11	&	26	&	20	\\
pn:\textsc{n}	&	-36\%	&	16	&	25	&	33	&	19	\\
subj:\textsc{n}	&	-33\%	&	6	&	9	&	10	&	7	\\
pp:\textsc{prep}	&	-30\%	&	14	&	20	&	19	&	23	\\
adv:\textsc{ptkneg}	&	-17\%	&	10	&	12	&	10	&	7	\\
det:\textsc{art}	&	-4\%	&	26	&	27	&	38	&	34	\\
\hline							
\it all		&	-48\%	&	222	&	429	&	493	&	488	\\
\hline
\end{tabular}
\caption{Main POS tags and dependency labels 
of words occurring in shifted blocks detected by HTER. NMT-vs-\uedin\ denotes the reduction of reordering errors in {\su} \textit{vs.} {\uedin} system.  Only word classes that were shifted 10 or more times in at least one system output are shown.}
 \label{tab:shiftClasses}
\end{table}
\setlength{\tabcolsep}{1pt}

The upper part of Table~\ref{tab:shiftClasses} shows that verbs are by far the most often misplaced word category in all PBMT systems -- an issue already known to affect standard phrase-based SMT between German and English \cite{Bisazza:13:WMT}. 
%Indeed, verbs can occupy very different positions between these languages. Moreover, 
Reordering is particularly difficult when translating \textit{into} German,  since the position of verbs in this language varies according to the clause type (\textit{e.g.} main \textit{vs.} subordinate).
Our results show that even syntax-informed PBMT does not solve this issue.
Using syntax at decoding time, as done by one of the systems combined within \uedin, appears to be a better strategy than using it for source pre-ordering, as done by the \hdu\ and \kit\ systems. 
%\hdu: dependency tree-based source preordering (D. Genzel, 2010 "Automatically learning source-side reordering rules for large scale machine translation")
%\kit: combined POS-based and constituency tree-based preordering: %T. Herrmann, J. Niehues, and A. Waibel, Combining Word Reordering Methods on different Linguistic Abstraction Levels for Statistical Machine Translation, 2013
%
However this only results in a moderate reduction of verb reordering errors (-12\% and \mbox{-25\%} \textit{vs.} \hdu\ and \kit\ respectively).
On the contrary, NMT reduces verb order errors by an impressive -70\% with respect to \uedin\ (-74\% and \mbox{-77\%} \textit{vs.} \hdu\ and \kit\ respectively)
despite being trained on raw parallel data without any syntactic annotation, nor explicit modeling of word reordering.
This result shows that the recurrent neural language model at the core of the NMT architecture is very successful at generating well-formed sentences even in languages with less predictable word order, like German (see examples in Table~\ref{table:examples}(a,b)).
NMT, though, gains notably less on nouns (\mbox{-47\%}), which is the second most often misplaced word category in \uedin. More insight on this is provided by the lower part of the table, where reordering errors are divided by their dependency label as well as POS tag. Here we see that 
%Looking at dependency labels, we find that
%According to dependency annotation,
order errors on nouns are notably reduced by NMT
when they act as syntactic objects (-65\% obja:\textsc{n}) but less when they act as preposition complements (-36\% pn:\textsc{n}) or subjects (-33\% subj:\textsc{n}).

The smallest NMT-\textit{vs}-\uedin\ gains are observed on prepositions (\mbox{-18\%} \textsc{prep}), negation particles (\mbox{-17\%} \textsc{ptkneg}) and articles (\mbox{-4\%} \textsc{art}). 
Manual inspection of a data sample reveals that
misplaced prepositions are often part of misplaced %CAMERA-READY
prepositional phrases acting, for instance, as temporal or instrumental adjuncts (e.g. \textit{`in my life'}, \textit{`with this video'}).
In these cases, the original MT output is overall understandable and grammatical, but does not conform to the order of German semantic arguments that is consistently preferred by post-editors (see example in Table~\ref{table:examples}(c)).
Articles, due to their commonness, are often misaligned by HTER and marked as \textit{shift} errors instead of being marked as two unrelated substitutions.
Finally, negation particles account for less than 1\% of the target tokens but play a key role in determining the sentence meaning. 
Looking closely at some error examples, we found that the correct placement of the German particle \textit{nicht} was determined by the focus of negation in the source sentence, which is difficult to detect in English.
For instance in Table~\ref{table:examples}(d) two interpretations are possible (\textit{`that did not work'} or \textit{`that worked, but not for systematic reasons'}), each resulting in a different, but equally grammatical, location of \textit{nicht}. 
In fact, negation-focus detection calls for a deep understanding of the sentence semantics, often requiring extra-sentential context \cite{Blanco:11}. 
When faced with this kind of translation decisions, NMT performs  as poorly as its competitors.

In summary, our fine-grained analysis confirms that NMT concentrates its word order improvements on important linguistic constituents and, specifically in English-German, is very close to solving the infamous problem of long-range verb reordering which so many PBMT approaches have only poorly managed to handle.
On the other hand, NMT still struggles with more subtle translation decisions depending, for instance, on the semantic ordering of adjunct prepositional phrases or on the focus of negation.

% CAMERA-READY: put back second verb reo example!!! DONE

\renewcommand{\arraystretch}{.9}
\begin{table*}[t]
\centering
\footnotesize
%\begin{tabular}{p{1.1cm} p{6.05cm} p{7.6cm}}
\begin{tabular}{l @{\ \ }l@{\ \ } l l}
  \toprule
  \multicolumn{3}{l}{\it Auxiliary-main verb construction [aux:\textsc{v}]:} \\ [.5mm]
& \textsc{src} & in this experiment , individuals \textbf{were shown} hundreds of hours of YouTube videos & \ \ \ \   \\ [.5mm]
\cmidrule{2-4}
& \textsc{hpb} & in diesem Experiment , Individuen \textbf{gezeigt wurden} Hunderte von Stunden YouTube-Videos &\multirow{2}{*}{\checkWrong}  \\
(a) & \textsc{pe} & in diesem Experiment \textbf{wurden} Individuen Hunderte von Stunden Youtube-Videos \textbf{gezeigt} & \\
%\cline{2-3}
%& \textsc{spb} & in diesem experiment , individuen hunderte stunden an youtube-videos , w√§hrend scans wurden ihre gehirne eine gro√üe bibliothek ihres gehirns reagiert , video-sequenzen gezeigt wurden \\
%& \textsc{pe} & in diesem experiment wurden individuen hunderte stunden an youtube-videos gezeigt , w√§hrend scans ihrer gehirne gemacht wurden , um eine gro√üe bibliothek ihres gehirns zu erstellen , das auf video-sequenzen reagiert \\
\cmidrule{2-4}
& \textsc{nmt} & in diesem Experiment \textbf{wurden} Individuen hunderte Stunden YouTube Videos \textbf{gezeigt} &\multirow{2}{*}{\checkRight}  \\
& \textsc{pe} & in diesem Experiment \textbf{wurden} Individuen hunderte Stunden YouTube Videos \textbf{gezeigt} & \\
 \midrule
 \multicolumn{3}{l}{\it Verb in subordinate (adjunct) clause [neb:\textsc{v}]:} \\ [.5mm]
& \textsc{src} & ... when coaches and managers and owners \textbf{look} at this information streaming ... \\ [.5mm]
\cmidrule{2-4}
& \textsc{pbsy} & ... wenn Trainer und Manager und Eigent\"{u}mer \textbf{betrachten} diese Information Streaming ... &\multirow{2}{*}{\checkWrong}  \\
(b) & \textsc{pe} & ... wenn Trainer und Manager und Eigent\"{u}mer dieses Informations-Streaming \textbf{betrachten} ... & \\
\cmidrule{2-4}
& \textsc{nmt} & ... wenn Trainer und Manager und Besitzer sich diese Informationen \textbf{anschauen} ... &\multirow{2}{*}{\checkRight}  \\
& \textsc{pe} & ... wenn Trainer und Manager und Besitzer sich diese Informationen \textbf{anschauen} ... & \\
 \midrule
%
%
%\multicolumn{3}{l}{\textit{Verb in object clause [objc:\textsc{v}]:}} & \\ [.5mm]
%& \textsc{src} & and so I claim that this \textbf{is} the same question as understanding the physical nature of intelligence & \\ [.5mm]
%\cline{2-3}
%& \textsc{synt} & und damit ich behaupten , dass dies die gleiche frage , wie das verst\"{a}ndnis der physikalischen natur der intelligenz \\
%(1) & \textsc{pe} & und so behaupte ich , dass dies die gleiche frage ist , wie das verst\"{a}ndnis der physikalischen natur der intelligenz \\
%\cmidrule{2-4}
%& \textsc{spb} & und ich behaupte , das \textbf{ist} die gleiche Frage wie das Verst\"{a}ndnis der physikalischen Natur der Intelligenz & \\
%(c) & \textsc{pe} & und so behaupte ich , dass dies die gleiche Frage \textbf{ist} , wie das Verst\"{a}ndnis der physikalischen Natur von Intelligenz & \\
%\cmidrule{2-4}
%& \textsc{nmt} & und so behaupte ich , dass dies die gleiche Frage \textbf{ist} , wie die physische Natur der Intelligenz zu verstehen  \\
%& \textsc{pe} & und so behaupte ich , dass dies die gleiche Frage \textbf{ist} , wie die physische Natur der Intelligenz zu verstehen  \\
 %\midrule

\multicolumn{3}{l}{\it Prepositional phrase [pp:PREP det:ART pn:N] acting as temporal adjunct:} \\ [.5mm]
& \textsc{src} & so like many of us , I 've lived in a few closets \textbf{in my life} \\ [.5mm]
\cline{2-4}
& \textsc{spb} & so wie viele von uns , ich habe in ein paar Schr\"{a}nke \textbf{in meinem Leben} gelebt &\multirow{2}{*}{\checkWrong}  \\
(c) & \textsc{pe} & so habe ich wie viele von uns \textbf{w\"{a}hrend meines Lebens} in einigen Verstecken gelebt & \\
\cmidrule{2-4}
& \textsc{nmt} & wie viele von uns habe ich in ein paar Schr\"{a}nke \textbf{in meinem Leben} gelebt &\multirow{2}{*}{\checkWrong}  \\
& \textsc{pe} & wie viele von uns habe ich \textbf{in meinem Leben} in ein paar Schr\"{a}nken gelebt & \\
 \midrule
 
\multicolumn{3}{l}{\it Negation particle [adv:\textsc{ptkneg}]:} \\ [.5mm]
& \textsc{src} & but I eventually came to the conclusion that that just did \textbf{not} work for systematic reasons \\ [.5mm]
\cmidrule{2-4}
& \textsc{hpb} & aber ich kam schlie√ülich zu dem Schluss , dass nur aus systematischen Gr\"{u}nden \textbf{nicht} funktionieren  &\multirow{2}{*}{\checkRight} \\
(d) & \textsc{pe} & aber ich kam schlie√ülich zu dem Schluss , dass es einfach aus systematischen Gr\"{u}nden \textbf{nicht} funktioniert &  \\
\cmidrule{2-4}
& \textsc{nmt} & aber letztendlich kam ich zu dem Schluss , dass das einfach \textbf{nicht} aus systematischen Gr\"{u}nden funktionierte &\multirow{2}{*}{\checkWrong} \\
& \textsc{pe} & ich musste aber einsehen , dass das aus systematischen Gr\"{u}nden \textbf{nicht} funktioniert  & \\
 \bottomrule
\end{tabular}
\caption{MT output and post-edit examples showing common types of reordering errors.}
\label{table:examples}
\end{table*}

\section{Conclusions}
We analysed the output of four state-of-the-art MT systems that participated in the English-to-German
task of the IWSLT 2015 evaluation campaign. %, which was focused on the translation of TED talks. 
Our selected runs were produced by three phrase-based MT systems and a neural MT system.
%a hierarchical phrase-based system (\hdu), a 
%phrase-based system (\kit), a combination of phrase-based and syntax-based system (\uedin), and finally 
%a neural MT system (\su).  
The analysis leveraged high quality post-edits of the MT outputs, which allowed 
us to profile systems with respect to reliable measures of post-editing effort and translation error types.
% general translation error types, and specific word order error sub-types. % translation errors related to word order.  

The outcomes of the analysis
confirm that NMT has significantly pushed ahead the state of the art, especially in a language pair
involving rich morphology prediction and significant word reordering.  
To summarize our findings:
\textit{(i)}  {\su} generates outputs that considerably lower the overall post-edit effort with respect to the best  PBMT 
system (-26\%); 
\textit{(ii)} {\su} outperforms PBMT systems on all sentence lengths, although its performance
degrades faster with the input length than its competitors; \textit{(iii)} {\su} seems to have an edge especially
on lexically rich texts; \textit{(iv)} {\su} output contains less morphology errors \mbox{(-19\%)}, less 
lexical errors \mbox{(-17\%)}, and substantially less word order errors \mbox{(-50\%)} than its closest competitor for each 
error type; \textit{(v)} concerning word order, {\su} shows an impressive improvement in the placement of 
verbs  \mbox{(-70\% errors)}.   

While {\su} proved superior to PBMT with respect to all 
error types that were investigated, our analysis also pointed out some aspects of {\su} that  
deserve further work, such as the handling of long sentences and the reordering of particular linguistic constituents requiring a deep semantic understanding of text.
%syntactic structures (negation, ...)
Machine translation is definitely not a solved problem, but the time is finally ripe to tackle its most intricate aspects.

\section*{Acknowledgments}
% Do not number the acknowledgment section.
FBK authors were supported by the CRACKER, QT21 and ModernMT projects, which received funding from the European Union's Horizon 2020 
%research and innovation 
programme under grants No. 645357, 645452 and 645487. AB was funded in part by the 
%Netherlands Organisation for Scientific Research
NWO under projects 639.022.213 and 612.001.218.

%LB and MC were supported by the CRACKER project, which received funding from the European Union’s Horizon 2020 research and innovation programme under grant no. 645357.
%MF was supported by the QT21 project, which received funding from the European Union’s Horizon 2020 research and innovation programme under grant agreement No. 645452.
%AB's work was funded in part by the Netherlands Organisation for Scientific Research (NWO) under project numbers 639.022.213 and 612.001.218.

%\bibliography{emnlp2016}
\bibliographystyle{emnlp2016}

\end{document}